\documentclass[runningheads]{llncs}
\usepackage{graphicx}
\usepackage{cite}
\usepackage{booktabs}
\usepackage[caption=false]{subfig}
\usepackage{amsmath}

\usepackage{hyperref}

\makeatletter
\newcommand{\printfnsymbol}[1]{%
  \textsuperscript{\@fnsymbol{#1}}%
}
\makeatother

\begin{document}

\setlength{\abovedisplayskip}{1pt}
\setlength{\belowdisplayskip}{1pt}
\setlength{\textfloatsep}{3pt}
%

\title{Hierarchical Self Attention Based Autoencoder for Open-Set Human Activity Recognition
}
\titlerunning{Hierarchical Self Attention Based Autoencoder for Open-Set HAR}
%
\author{M Tanjid Hasan Tonmoy\thanks{Equal contribution} \and
Saif Mahmud\printfnsymbol{1} \and
A K M Mahbubur Rahman \and
M Ashraful Amin \and
Amin Ahsan Ali}
\authorrunning{M.T.H. Tonmoy and S. Mahmud et al.}
%
\institute{Independent University Bangladesh, Dhaka, Bangladesh\\
\email{\{2015-116-770, 2015-116-815\}@student.cse.du.ac.bd}\\
\email{\{akmmrahman, aminmdashraful, aminali\}@iub.edu.bd}}
\maketitle              
\begin{abstract}
Wearable sensor based human activity recognition is a challenging problem due to difficulty in modeling spatial and temporal dependencies of sensor signals. Recognition models in closed-set assumption are forced to yield members of known activity classes as prediction. However, activity recognition models can encounter an unseen activity due to body-worn sensor malfunction or disability of the subject performing the activities. This problem can be addressed through modeling solution according to the assumption of open-set recognition. Hence, the proposed self attention based approach combines data hierarchically from different sensor placements across time to classify closed-set activities and it obtains notable performance improvement over state-of-the-art models on five publicly available datasets. The decoder in this autoencoder architecture incorporates self-attention based feature representations from encoder to detect unseen activity classes in open-set recognition setting. Furthermore, attention maps generated by the hierarchical model demonstrate explainable selection of features in activity recognition. We conduct extensive leave one subject out validation experiments that indicate significantly improved robustness to noise and subject specific variability in body-worn sensor signals. The source code is available at: \href{https://github.com/saif-mahmud/hierarchical-attention-HAR}{github.com/saif-mahmud/hierarchical-attention-HAR}

\end{abstract}

\section{Introduction}
Automated Human Activity Recognition (HAR) has a pivotal role in mobile health, physical activity monitoring, and rehabilitation. Body-worn sensor based HAR can broadly be defined as classification of human physical activity based on signals from multitude of wearable sensors worn at different body locations. Human physical activities include activities of daily living as well as complex activities comprised of multiple simpler micro-activities. Increasing usage of smart handheld devices with multi-modal sensors has paved the way to deploy HAR system in applications of elderly activity monitoring, physiotherapy exercise evaluation, and smart home solutions.

HAR techniques rely on spatial information and temporal dynamics of physical activity captured from heterogeneous sensors placed at different human body locations. Activities involve different dominant body parts and thus hierarchical fusion of sensor signals from different sensor placements is able to capture salient information required for the task. Most human activities can be viewed as a session comprising a number of short time windows containing low level actions. Hierarchical fusion of temporal information is able to take this phenomenon into account. Further, HAR systems in real world scenarios are likely to encounter samples different from training classes. An optimal framework should be able to distinguish them from known classes with explainable visualizations.


Though initial works relied on domain knowledge and heuristic based statistical feature representation \cite{plotz2011feature} for activity recognition, recent end-to-end deep learning models utilize convolutional \cite{hammerla2016deep} and recurrent \cite{convLSTM:HAR} architectures for learning representations. Attention mechanism, described as weighted average of feature representations, is adopted to model HAR task like other sequence modeling problem such as NLP in recent works \cite{murahari2018attention, ContAttn:HAR, attnsense:HAR, uniqueAttn:HAR, ecai2020}. However, such approaches do not utilize hierarchical modelling of spatio-temporal information. Moreover, these approaches follow conventional training method under closed set assumption where unknown samples are forced to be recognised as one of the prior known classes. 


Considering the aforementioned requirement towards hierarchical fusion of spatio-temporal features, the methodological approach taken in this paper incorporates hierarchical aggregation of sensor signals from different placements across time to construct representation for a specific window. Feature representations from different windows within the same session are aggregated to yield representation for that session. In the case of predicting label for specific window, we utilize session information guided window feature representation instead. The proposed approach models HAR under open set assumption where test samples are identified as seen or unseen and labeled as one of the known classes from training set simultaneously. Such capabilities are highly desirable in the scenario of a subject performing an activity in a completely unexpected way e.g. doing rehabilitation exercise incorrectly due to physical limitations or malfunctioning of sensor devices. In this regard, we have designed autoencoder architecture along with hierarchical encoder to model the distribution of the known activity classes where unseen activities are supposed to yield higher reconstruction loss. Explainable feature attention maps are obtained from hierarchical self-attention layers to demonstrate dominant sensor placement and temporal importance within session to classify specific physical activity. 

We conduct extensive experiments on five publicly available benchmark activity recognition datasets: PAMAP2, Opportunity, USC-HAD, Daphnet and Skoda. Our model outperforms prior methods in several datasets. Furthermore, we evaluate the robustness of the model to noise and subject-specific variability through leave-one-subject-out validation experiments. Moreover, we evaluate open-set recognition performance and generate feature attention maps to demonstrate the activity distinguishing characteristics in the learned representation. In brief, the key contributions of our work are listed below:
\begin{enumerate}
    \item Proposed hierarchical self attention encoder models spatial and temporal information of raw sensor signals in learned representations which are used for closed-set classification as well as detection of unseen activity class with decoder part of the autoencoder network in open-set problem definition 
	\item Interpretable visualization of feature attention maps indicate fusion of causal and coherent features for activity recognition
	\item Our extensive experiments achieve superior performance in several benchmark datasets and demonstrate robustness to subject-specific variability in sensor readings
\end{enumerate}


\section{Related Work}
\label{related_works}

\textbf{Wearable Sensor Based HAR} The earlier research works on HAR mostly relied on hand-crafted statistical or distribution-based \cite{kwon2018adding, qian2019novel} features that have been designed based on domain expertise \cite{onFeature_ISWC2019:HAR}. However, recent works for wearable-based HAR have mostly focused on end to end deep learning systems for modeling effective feature representation. In that regard, various forms of convolutional, recurrent and hybrid architectures such as \cite{hammerla2016deep, guan2017ensembles, convLSTM:HAR} were proposed and demonstrated varying levels of success in recognition of the activities under consideration. In recent years, the incorporation of attention mechanism with deep learning-based architectures \cite{murahari2018attention, ContAttn:HAR, uniqueAttn:HAR, attnsense:HAR} have demonstrated significant performance improvement. However, most of these works \cite{ijcai2018:complexHAR, peng2018:complexHAR} do not rely on hierarchical modeling of spatio-temporal information from wearable sensors.

\textbf{Self Attention Architecture} Recently, self-attention \cite{Transformer_NIPS2017} based models have emerged as a popular alternative to recurrent networks for various NLP tasks and has also been proposed for HAR \cite{ecai2020}. Using self-attention in a hierarchical manner has been proposed for various tasks such as classifying text documents \cite{hsa-cancer-pathology:NLP}, generating recommendations \cite{he2019hierarchical} etc. in order to break up the task into relevant hierarchical parts. However, no such work exists for wearable sensor data to the best of our knowledge though such hierarchy allows for intuitive representation of complex human activities.

\textbf{Interpretability and Open-Set Recognition} The data from wearable sensor devices are usually high dimensional involving different body placements over some time duration. Furthermore, most of the deep-learning based models used for classification of such data offer little to zero interpretability towards the predicted outcome. Some progress has been made in this regard for video based action recognition task \cite{meng2019interpretable,ramakrishnan2019identifying}. Although attention-based models for wearable HAR offer more interpretability using the attention-scores, it is still scarce in HAR community. With regards to HAR systems, it is often useful to be able to identify previously unseen activities. Class conditioned \cite{oza2019c2ae} or variational \cite{vasilev2018qspace, an2015variational} autoencoder for unseen sample detection has been proposed for image data. Although unseen activity recognition for skeleton data \cite{novelty-skeleton-data} \& smart-home environment \cite{novelty-smart-home-1,novelty-smart-home-2} have been proposed, autoencoder architectures for unseen sample detection in wearable based HAR are very scarce.

\section{Proposed Method}
\subsection{Task Definition}
We assume that $S=\{S_1, S_2, S_3, ...\}$ is the set of sensors placed at different locations on the body of human subjects. Generally, an Inertial Measurement Unit (IMU) or smart-device contains sensors and records data at a sampling rate of $f$ Hz from multiple axes (e.g. tri-axial accelerometer or gyroscope yields signal along $x$, $y$ and $z$-axis). Therefore, there will be record of $m = |S| * \sum_i (a_i)$ signals at any particular time-step where $a_i = $ number of axes at $S_i$ . In a dataset containing sensor signal recording of $n$ time-steps, the readings along time is represented as a multidimensional time-series $X$ as in (\ref{eqn:sensor-timeseries}) and reading at particular time-step $x_k$ can be represented as in (\ref{eqn:sensor-timestep}). The human activity recognition problem can be defined as the task to detect physical activity class labels given the multidimensional time-series of sensor signals $X$ of particular duration.
\begin{equation}
\label{eqn:sensor-timeseries}
X = [x_1, x_2, ... , x_k, ... , x_n]
\end{equation}
\begin{equation}
\label{eqn:sensor-timestep}
x_k = [x_{k1}, x_{k2}, ..., x_{kj}, ..., x_{km}]^T
\end{equation}
We propose to represent sensor signals hierarchically as an activity session composed of windows representing short segments within the sequence. On the other hand, a window is composed of a fixed number of data-points representing the sensor signal at the corresponding time-stamps. We use the proposed hierarchical encoder to learn representation for a session which is used for both classification and open-set detection. The different components of the model are described in the following subsections. 

\subsection{Hierarchical Self Attention Encoder}
The proposed model incorporates two distinct types of hierarchy - temporal and body location-based. 
These hierarchies are implemented by the Hierarchical Window Encoder (HWE) and Session Encoder (SE), respectively. Self attention is the core element in both of the aforementioned components and is used in two ways within the components. We refer to them as Modular Self Attention and Aggregator Self Attention, respectively.

\textbf{Modular Self Attention:}
Modular self attention consists of $N$ identical blocks of multi-headed self attention and position-wise feed forward layers. For each time-step, three linear transformations referred to as key ($K$), query ($Q$), and value ($V$) are learned. Attention score is obtained by applying softmax function on the scaled dot products of queries and keys and is used to get a weighted version of the values. This operation is performed in matrix form as defined in (\ref{eqn:sa}).
\begin{equation}
\label{eqn:sa}
f_{sa}(Q,K,V) = \textnormal{softmax}( \frac{QK^T}{\sqrt{d_k}})V
\end{equation} 
Here, $Q = X W_Q, K = X W_K, V = X W_V$ where $W_Q, W_K, W_V$ are weight matrices; X is the input and $K\in{\rm I\!R}^{t \times d_k}, Q\in{\rm I\!R}^{t \times d_k}\textnormal{ and }V\in{\rm I\!R}^{t \times d_v}$.
Furthermore, multi-head self attention is defined in (\ref{eqn:sa-multi-head}) where each attention head uses different $W_Q, W_K, W_V$ and the output from different attention heads are combined according to (\ref{eqn:sa-multi-head}).
\begin{equation}
\label{eqn:sa-multi-head}
\mathbf{f}_{mhsa} (X) = concat (h^{(1)},...., h^{(n)}) \cdot \mathbf{W}_o
\end{equation}
where, $h^{(j)} = f_{sa}(W_Q^{(j)}X, W_K^{(j)}X, W_V^{(j)}X)$. Position-wise feed forward refers to identical fully connected feed forward network composed of two feed forward layers with ReLU activation in between applied independently to each time-step. In addition, layer normalization is used after self-attention and position-wise feed forward layers and the aforementioned layers contain residual connections.

\textbf{Aggregator Self Attention:}
In order to obtain an aggregate representation from all the time-steps in the input sequence, we use Aggregator Self Attention. The primary difference between Modular and Aggregator blocks is in the learned linear representations used in (\ref{eqn:agr-sa}). The construction of query and value are the same as the former. The key matrix $K_a$ in (\ref{eqn:agr-sa}) is initialized randomly and learned during optimization with the rest of the parameters.
\begin{equation}
\label{eqn:agr-sa}
f_{agr}(Q_a,V_a) = \textnormal{softmax}( \frac{Q_aK_a^T}{\sqrt{d_{ka}}})V_a
\end{equation} 
Where, $Q_a = X W_{aQ}, V = X W_{aV}; W_{aQ}, W_{aV}$ are weight matrices; $X$ is the input to the layer and $Q_a\in{\rm I\!R}^{t \times d_{ka}}, V_a\in{\rm I\!R}^{t \times d_{va}}, K_a\in{\rm I\!R}^{1 \times d_{ka}}$.
In contrast to Modular Self Attention, the position-wise feed forward layer is applied both before and after the self attention operation. Moreover, we use single headed attention in this block in order to simplify the use of the attention scores for interpretability. 

\begin{figure}[t!]
\centering
\subfloat[Hierarchical Self Attention Encoder\label{fig:hsa-encoder}]{%
  \includegraphics[width=0.45\textwidth, height=7.0cm]{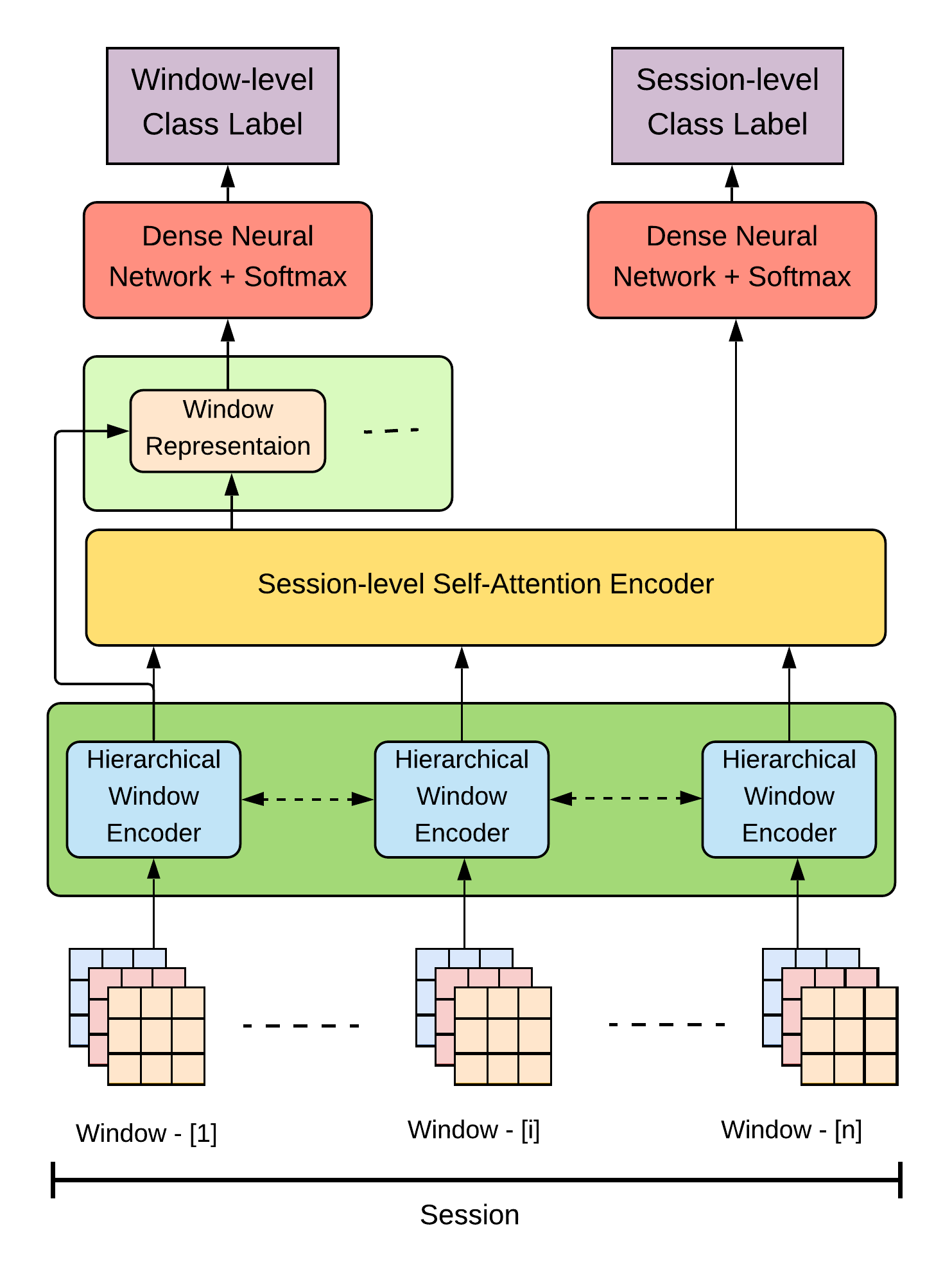}%
}\hfil
\subfloat[Hierarchical Window Encoder\label{fig:hwe}]{%
  \includegraphics[width=0.43\textwidth, height=7.0cm]{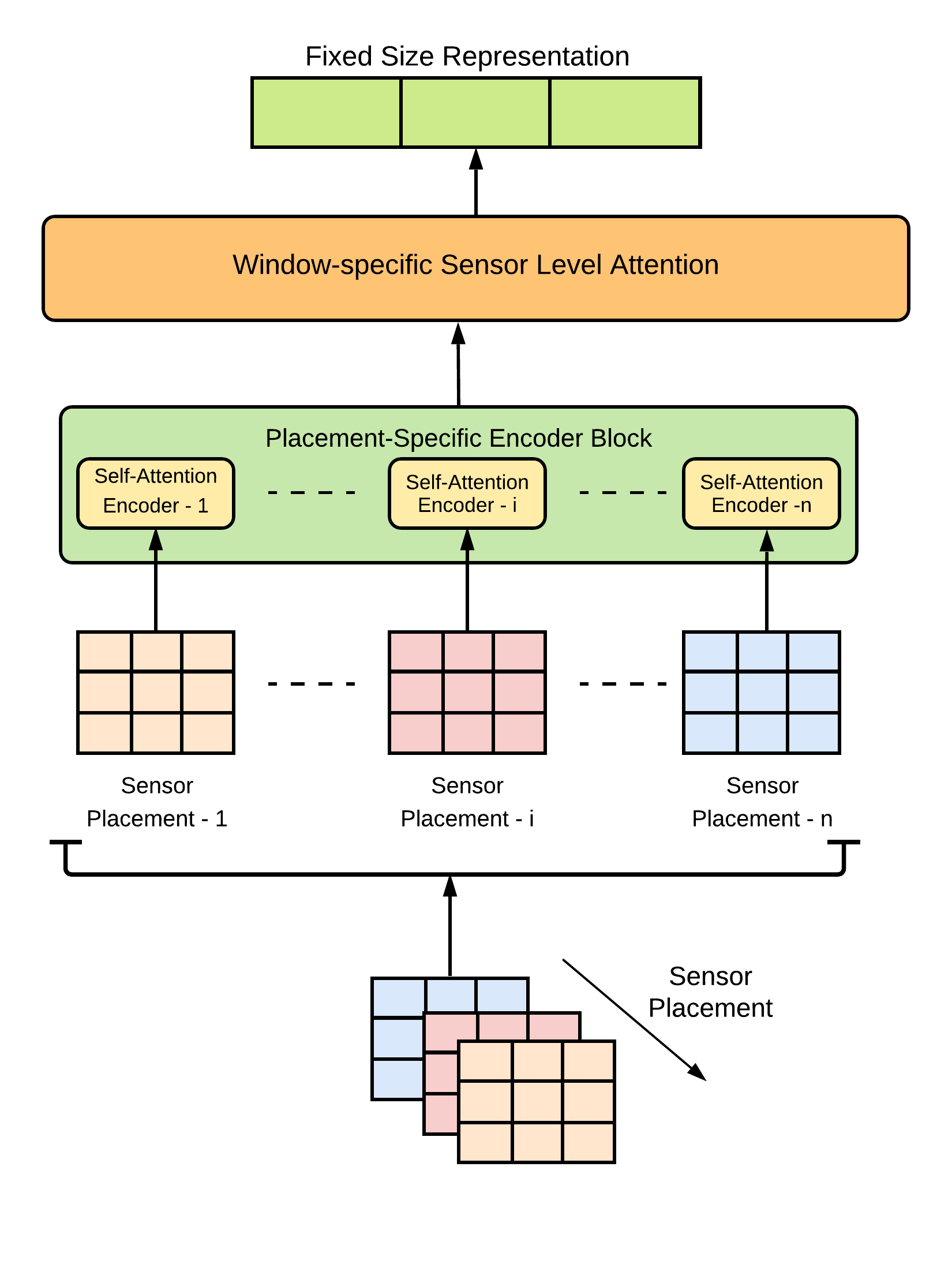}%
}
\caption{Overview of the (a) \textbf{Hierarchical Self Attention Encoder}, consisting of Hierarchical window encoders and session encoder, is used to obtain representation for classification and open set detection
\newline
(b) \textbf{Hierarchical Window Encoder}, where sensor signals from different body locations within short time span are separately transformed and fused later using self attention
}
\end{figure}

\textbf{Hierarchical Window Encoder (HWE)} consists of $m$ modular self attention blocks and an aggregator self attention block where $m$ is the  number of sensor placements. 
First, the values from all sensor modalities placed in a body location are combined using $1$-D convolution over single time-step to create a $d_{model}$ sized vector. The position information is incorporated by adding positional encoding based on sine and cosine function. Afterwards, the sequence from each sensor placement goes through modular self attention block. Finally, the transformed time-steps are concatenated along the temporal axis and aggregator self attention is used to obtain a representation for the window as defined in (\ref{eqn:window-encoder-sa}) and (\ref{eqn:window-encoder-agr}).
\begin{equation}
\label{eqn:window-encoder-sa}
 Z_{window} = concat (f_{mhsa}(X_w^{(i)})\textnormal{, ... ,} f_{mhsa}(X_w^{(m)}))
\end{equation}
\begin{equation}
\label{eqn:window-encoder-agr}
\tilde{Z}_{window} = f_{agr} (Z_{window})
\end{equation}
\textbf{Session Encoder (SE)} comprises $n$ identical HWEs where $n$ is the number of windows within the session. All of the HWE within the session have shared parameters. The input to the session encoder is the output from $n$ temporally ordered window encoders where $n$ is the number of windows within the session. Similar to the HWE, the input goes through Modular Self Attention and Aggregator Self Attention as defined in (\ref{eqn:session-encoder-sa}) and (\ref{eqn:session-encoder-agr}) to obtain a representation for the session. 
\begin{equation}
\label{eqn:session-encoder-sa}
Z_{session} = f_{mhsa}(X_s)
\end{equation}
\begin{equation}
\label{eqn:session-encoder-agr}
\tilde{Z}_{session} = f_{agr} (Z_{session})
\end{equation}
\textbf{Window and Session Classification:}
For session-level classification, the output from SE is passed to dense and softmax layers to obtain the class label. However, for widow-level classification we concatenate the output from SE with each window representation and pass that to dense and softmax layers. Therefore, we utilize the hierarchical structure to augment the window representation with session information to guide the window-level classification. 

\subsection{Open Set Human Activity Recognition}
Autoencoder, constructed upon the proposed hierarchical self-attention encoder, models the relationship between random variable $z$ representing low-dimensional latent space and random variable $x$ denoting learned representation vector to be reconstructed. We have designed the decoder as multi-layer feed-forward neural network estimating the approximation of distribution $p_\theta(x|z)$ where $\theta$ is the learned decoder parameters. On the other hand, the encoder is trained to model the posterior distribution $q_\phi(z|x)$ where $\phi$ indicates encoder network parameters.

\begin{figure}[t]
	\centering
	\includegraphics[width=.8\linewidth]{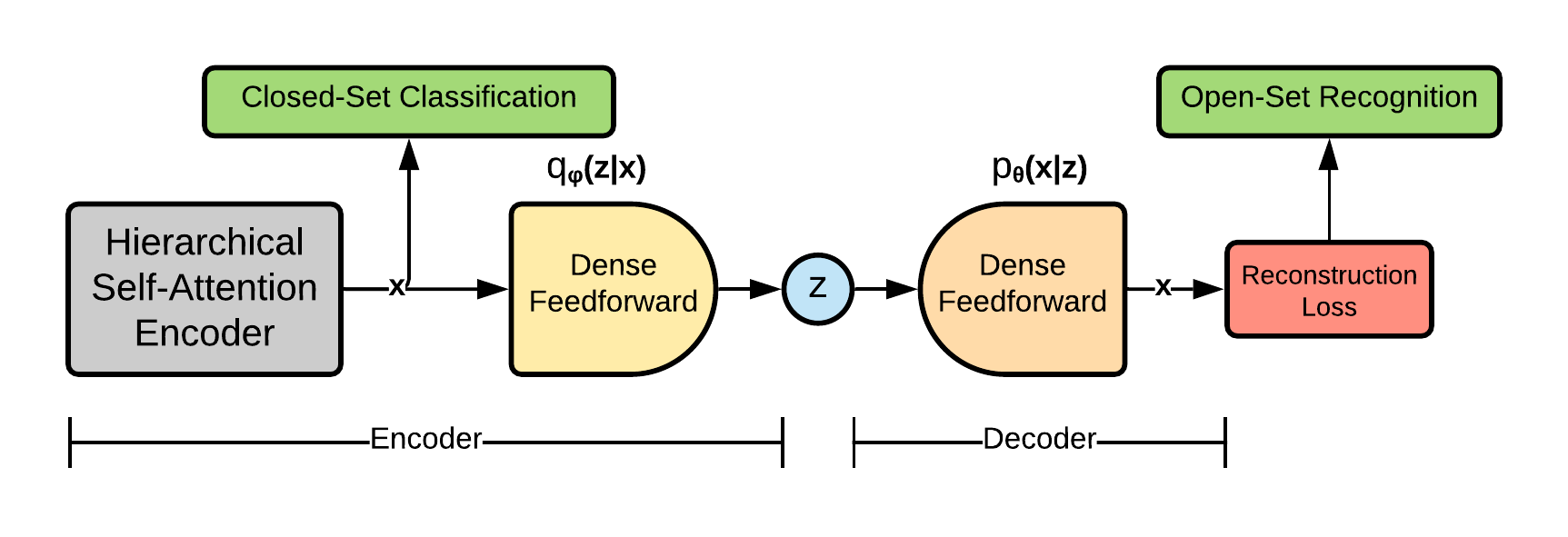}
	\caption{Overview of the autoencoder architecture where representations from hierarchical self attention encoder is utilized for closed-set classification and reconstructed with decoder for open-set recognition}
	\label{fig:vae}
\end{figure}

As illustrated in Fig. \ref{fig:vae}, the objective of proposed autoencoder is to approximate the intractable true posterior $p_\theta(z|x)$ with $q_\phi(z|x)$. The approximation depends on the network parameters and they are tuned based on reconstruction loss and Kullback–Leibler (KL) divergence $D_{KL}(q_\phi(z|x) || p_\theta(z|x))$. As the KL divergence cannot be computed directly from feature representation $x$ and latent space representation $z$, the loss is minimized through maximizing summation of Evidence Lower Bound (ELBO). Therefore, the loss of autoencoder is computed as, $\mathcal{L}_{AE} = - \sum_{i} ELBO_i$, where $ELBO_i$ is defined as below in (\ref{eqn:elbo}):
\begin{equation}
\label{eqn:elbo}
ELBO_i = {\rm I\!E}_{q_\phi(z|x_i)}[log(p_\theta(z|x_i))] - D_{KL}(q_\phi(z|x_i) || p_\theta(z))
\end{equation}

Here, $ELBO_i$ is the Evidence Lower Bound on the marginal likelihood of the $i$-th learned representation, $x_i$. $p_\theta(z)$ indicates prior probability and modeled as unit Gaussian. The expected value defined in the first term indicates reconstruction loss of learned representations. It is assumed according to the characteristics of autoencoder that known activity classes will demonstrate lower reconstruction loss whereas unseen or novel ones should yield higher. The novel activities are detected based on reconstruction loss threshold which is tuned as hyperparameter. The threshold is set from the range $\mu( \mathcal{L}_{known}) - \alpha \cdot \sigma(\mathcal{L}_{known})$ where $\mathcal{L}_{known}$ is the reconstruction loss of autoencoder on training data containing known activity classes and $\alpha \in [0.0, 0.50]$.

\section{Experiments}
\textbf{Datasets: }We use five publicly available benchmark datasets - PAMAP2~\cite{Reiss:2012:INB:2357489.2358027}, Opportunity~\cite{opportunity-dataset}, USC-HAD \cite{usc-had_paper}, Daphnet \cite{daphnetdataset} and Skoda \cite{skoda-paper} for our experiments. A summary of the datasets used is presented in Table~\ref{table:dataset}.

\begin{table}[b]
\caption{Summary of the Datasets used in experiments. For Skoda, we use a 10\% split for test and validation since it contains data from a single subject. For the \textit{Sensor Used} column, A $=$ Accelerometer, G $=$ Gyroscope, M $=$ Magnetometer}
\label{table:dataset}
\centering
\resizebox{\columnwidth}{!}{
\begin{tabular}{@{}l||c|c|c|c|c|c@{}}
\toprule
\multicolumn{1}{c||}{\textbf{Dataset}} & \textbf{\begin{tabular}[c]{@{}c@{}}Sampling\\ Rate\end{tabular}} & \textbf{\begin{tabular}[c]{@{}c@{}}No. of\\ Activity\end{tabular}} & \textbf{\begin{tabular}[c]{@{}c@{}}No. of\\ Subject\end{tabular}} & \textbf{\begin{tabular}[c]{@{}c@{}}Validation\\ Subject ID\end{tabular}} & \textbf{\begin{tabular}[c]{@{}c@{}}Test\\ Subject ID\end{tabular}} & \textbf{\begin{tabular}[c]{@{}c@{}}Sensor\\ Used\end{tabular}} \\ \midrule
PAMAP2                               & 100 Hz                                                           & 12                                                                 & 9                                                                 & 105                                                                   & 106                                                             & A, G                                                           \\
Opportunity                          & 30 Hz                                                            & 18                                                                 & 4                                                                 & 1 (run 2)                                                             & 2, 3 (run 4, 5)                                                   & A, G, M                                                        \\
USC-HAD                              & 100 Hz                                                           & 12                                                                 & 14                                                                & 11, 12                                                                & 13, 14                                                          & A, G                                                           \\
Daphnet                              & 64 Hz                                                            & 2                                                                  & 10                                                                & 9                                                                     & 2                                                               & A                                                              \\
Skoda                                & 98 Hz                                                            & 11                                                                 & 1                                                                & 1 (10\%)                                                              & 1 (10\%)                                                        & A                                                              \\ \bottomrule
\end{tabular}}
\end{table}


\textbf{Construction of Activity Window \& Session:}
Activity sessions are constructed using overlapping sliding-window across the temporal axis. Each activity session comprises a number of non-overlapping short activity segments which we refer to as windows. 

\textbf{Open-set Experiment:} We randomly hold out the data for a fraction of the classes (22\% and 27\% of the classes in case of Oppotunity \& Skoda, 33.33\% for rest) as part of the open-set and include the benchmark test data as defined in Table~\ref{table:dataset} to construct the test set for evaluation.
We train the model with the remaining data and report the cross validation results.

\textbf{Training and Hyperparameters:}
We implement the proposed model in Tensorflow and train on eight Tesla K80 GPUs. We use Adam optimizer with learning rate set to $10^{-3}$ with momentum $\beta_1 = 0.9$ and $\beta_2 = 0.999$ and weight decay $\epsilon = 10^{-7}$. The number of identical blocks and head for multi-headed self-attention is set to $N = 2$ and $n = 4$ respectively for our experimental setup. The dropout applied to placement specific encoder block and the size of the representation vector learned from each session is configured to $0.2$ and $d_{model} = 64$ respectively.

\section{Results and Discussion}

\textbf{Evaluation Metric:}
For the evaluation of activity recognition performance, we use the macro average F1-score defined in (\ref{eqn:f1score}) as metric where $|C|$ and $ i = 1, ..., C$ in (\ref{eqn:f1score}) indicate number of classes and the set of classes respectively.
\begin{equation}
\label{eqn:f1score}
\textnormal{Macro F1} = \frac{1}{|C|} *\sum_{i = 1}^{C} \frac{2 * Precision_i * Recall_i}{Precision_i + Recall_i}
\end{equation}

\subsubsection{Baselines and Performance Comparison}
We compare the proposed method with a number of baselines which includes most of the prominent feature-based deep learning methods for HAR as well as the recent state-of-the-art models. In particular, we compare our approach with recurrent, convolutional, hybrid and attention-based models. Recurrent network based baselines include LSTM and b-LSTM \cite{hammerla2016deep}. For convolutional baselines we compare with simple CNN as well as convolutional autoencoder. Hybrid baselines include DeepConvLSTM (4 CNN and 2 LSTM layers). Attention based baselines include attention augmented DeepConvLSTM as well as SADeepSense \cite{yao2019sadeepsense} (CNN, GRU and sensor \& temporal self-attention modules) and AttnSense \cite{attnsense:HAR} (attention based modality fusion subnet and GRU subnet). We also compare with self-attention based transformer classifier\cite{ecai2020} which does not use any hierarchical modelling. 
\begin{table}[t!]
\centering
\caption{Performance comparison of the proposed method with baselines in terms of window-wise results on benchmark test set}
\label{table:performance-comparison}
\resizebox{\columnwidth}{!}{
\begin{tabular}{@{}lccccc@{}}
\toprule
\textbf{Methods}                  & \textbf{PAMAP2} & \textbf{Opportunity} & \textbf{USC-HAD} & \textbf{Daphnet} & \textbf{Skoda} \\ \midrule
LSTM (2014)                       & 0.75            & 0.63                 & 0.38             & 0.68             & 0.89           \\
CNN (2015)                        & 0.82            & 0.59                 & 0.41             & 0.59             & 0.85           \\
b-LSTM \cite{hammerla2016deep} (2016)                     & 0.84            & 0.68                 & 0.39             & 0.74             & 0.91           \\
DeepConvLSTM  \cite{convLSTM:HAR} (2016)               & 0.75            & 0.67                 & 0.38             & 0.84             & 0.91           \\
Conv AE \cite{onFeature_ISWC2019:HAR} (2017)                    & 0.80            & \textbf{0.72}        & 0.46             & 0.73             & 0.79           \\
DeepConvLSTM + Attn \cite{murahari2018attention} (2018)        & 0.88            & 0.71                 & 0.51             & 0.76             & 0.91           \\
SADeepSense \cite{yao2019sadeepsense} (2019)                & 0.66            & 0.66                 & 0.49             & 0.80             & 0.90           \\
AttnSense \cite{attnsense:HAR} (2019)                 & 0.89            & 0.66                 & 0.49             & 0.80             & 0.93           \\
Transformer Encoder \cite{ecai2020} (2020)        & 0.96            & 0.67                 & 0.55             & 0.82             & 0.93           \\
\textbf{Proposed HSA Autoencoder} & \textbf{0.99}   & 0.68                 & \textbf{0.55}    & \textbf{0.85}    & \textbf{0.95}  \\ \bottomrule
\end{tabular}}
\end{table}

\textbf{Performance on Benchmark Test Set:}
Table~\ref{table:performance-comparison} shows that the proposed model outperforms the baseline methods for all of the datasets except Opportunity in terms of window-based results. Specifically, the proposed method outperforms the other methods for PAMAP2, Skoda and Daphnet dataset. On USC-HAD dataset, the performance of the proposed model is on par with the transformer encoder which can be explained by the fact that the dataset contains sensor readings from only one body location (waist) thus not being able to take advantage of sensor location hierarchy. With regards to the Opportunity dataset, some of the mid-level gestures are very short (e.g less than one second) for which the hierarchical model does not improve much on the existing results. However, the advantage of the proposed hierarchy becomes apparent when we consider the performance on longer and more complex activities. In particular, for the recognition of $5$ high level complex activities in the Opportunity dataset, we observe better performance compared to the others (proposed model obtains macro-F1 of $0.91$ compared with $0.71$, $0.73$, $0.791$, $0.838$ for CNN, LSTM, DeepConvLSTM and AROMA \cite{peng2018:complexHAR} respectively).
Therefore, the proposed hierarchical method not only produces better performance in case of longer complex activities (session-wise) but also improves the recognition of shorter activities (window-wise).
\begin{table}[t!]
\caption{Performance Evaluation in LOSO Experiment}
\label{table:loso}
\centering
\begin{tabular}{@{}c|c|c|c|c@{}}
\toprule
\textbf{Dataset} & \textbf{Proposed Model} & \textbf{\begin{tabular}[c]{@{}c@{}}Transformer Encoder\end{tabular}} & \textbf{DeepConvLSTM} & \textbf{Conv AE} \\ \midrule
PAMAP2                & \textbf{0.94}           & 0.92                                                                   & 0.61                  & 0.48                     \\
Opportunity           & 0.43                    & 0.42                                                                   & \textbf{0.44}         & 0.42                     \\
USC-HAD               & \textbf{0.68}           & 0.60                                                                   & 0.59                  & 0.63                     \\
Daphnet               & \textbf{0.72}           & 0.71                                                                   & 0.69                  & 0.67                     \\ \bottomrule
\end{tabular}
\end{table}

\textbf{Performance on Leave-One-Subject-Out (LOSO) Experiment:}
In order to demonstrate the proposed hierarchical model's robustness  to subject specific variability in sensor reading, we perform leave-one-subject-out (LOSO) validation experiments. In this regard, we hold data of one subject out for evaluation. Table~\ref{table:loso} presents macro F1 score of LOSO experiments on four datasets (Skoda contains single subject) used for experiments. As can be seen from the table, the model demonstrates better performance on LOSO experiments compared to benchmark test data while other models suffer from subject specific variable sensor reading patterns for the same activity.

\begin{figure}[t!]
	\centering
	\includegraphics[width=.6\linewidth, height=6.0cm,scale=0.75]{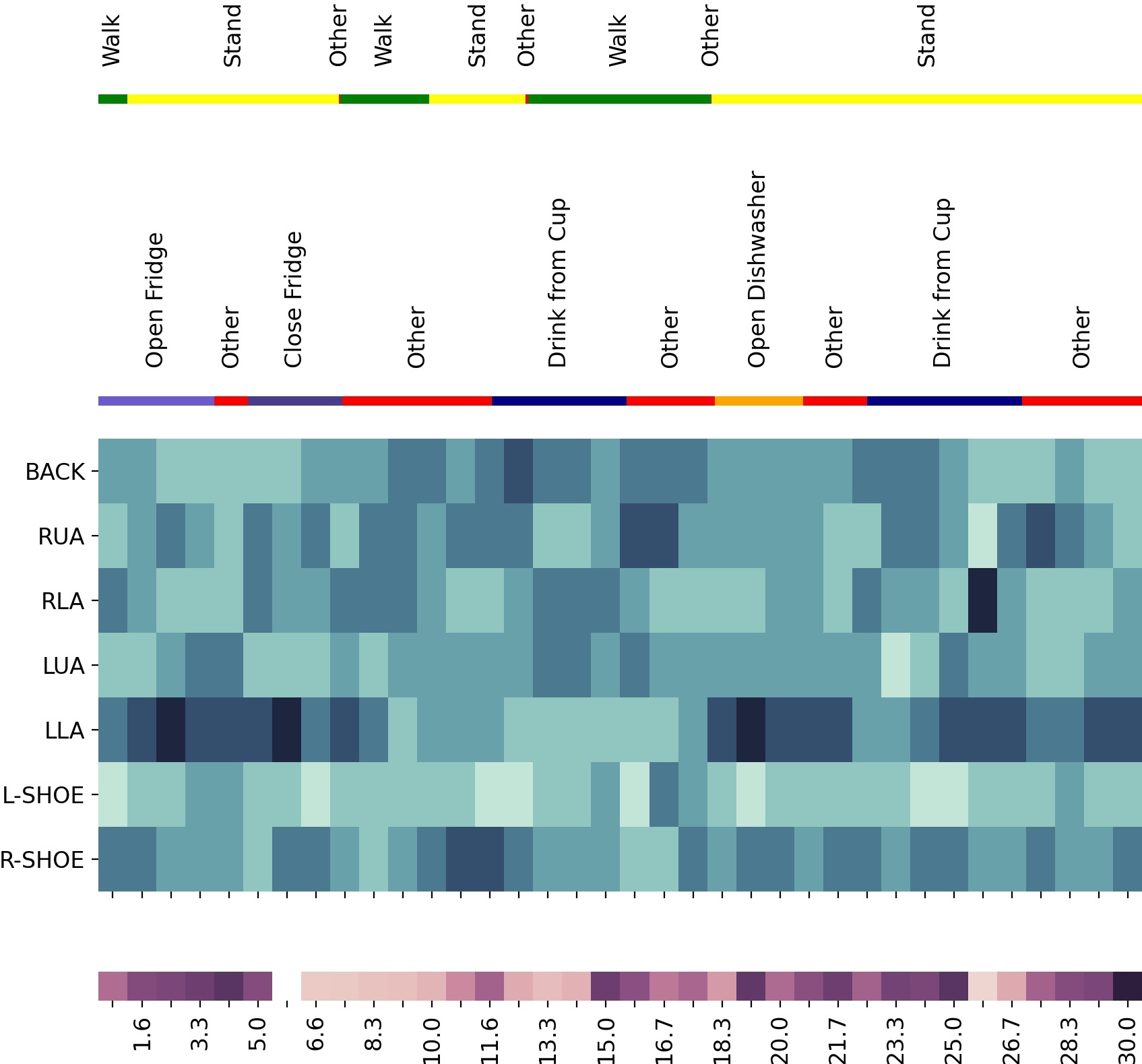}
	\caption{Attention map for activity `Cleanup' from Opportunity dataset comprising locomotion and mid-level gestures (top two rows - plotted using ground truth annotation), bottom  $x$-axis shows temporal attention weight and $y$-axis indicates body locations of sensors [(L = Left, R = Right), (L = Lower, U = Upper) \& A = Arm]  where darker color indicates higher attention score}
	\label{fig:opp-attentionmap-cleanup}
\end{figure}

\textbf{Attention Maps for Interpretability:}
We can obtain temporal and sensor-placement specific attention maps based on the attention scores obtained from SE and HWE respectively. The attention maps are useful for understanding the predictions made by the model. With regards to temporal attention maps, a snapshot of which time frames were of more importance for the prediction is useful for understanding both the model output and activity itself in case of unseen activities. Moreover, such attention maps demonstrate good correspondence with mid-level or micro activities that comprise the action. One such example is illustrated in Fig.~\ref{fig:opp-attentionmap-cleanup} for a complex activity from the Opportunity dataset using the annotated mid level actions and locomotion. It is evident that more emphasis is given on the relevant actions for recognition of the activity. Furthermore, the sensor-placement based attention maps provide a  finer granularity of information regarding which placements played more prominent roles at different times which is in line with the intuitive understanding that distinct micro activities may be dominated by different body parts.

\textbf{Performance on Open set HAR:}
Furthermore, the proposed model also produces noteworthy results in the case of open set recognition as shown in Table~\ref{table:openset-result}. The proposed auto-encoder obtains good performance in terms of accuracy and macro F1-score on PAMAP2, Opportunity and USC HAD dataset indicating the capability to distinguish between the activities belonging to the known \& unknown classes. With regards to Daphnet dataset, the scores are lower compared to rest since the unknown class includes transition activities in between the two known classes resulting in similar distribution for known and unknown activities.

\begin{table}[t!]
\centering
\caption{Open Set Activity Detection Performance}
\label{table:openset-result}
\begin{tabular}{@{}ccccc@{}}
\toprule
\textbf{Dataset} & \textbf{\begin{tabular}[c]{@{}c@{}}Total Number\\ of Classes\end{tabular}} & \textbf{\begin{tabular}[c]{@{}c@{}}Number of\\ Novel Classes\end{tabular}} & \textbf{Accuracy} & \textbf{\begin{tabular}[c]{@{}c@{}}Macro\\  F1 Score\end{tabular}} \\ \midrule
PAMAP2 & 12 & 4 & 0.85 & 0.69 \\
Opportunity & 18 & 4 & 0.75 & 0.58 \\
USC HAD & 12 & 3 & 0.61 & 0.52 \\
Skoda & 11 & 3 & 0.55 & 0.44 \\
Daphnet & 3 & 1 & 0.42 & 0.39 \\ \bottomrule
\end{tabular}
\end{table}

\section{Conclusion}
The aim of this work was to design self-attention based model with hierarchical fusion of spatial and temporal features. Our extensive experiments confirmed that hierarchical aggregation leads to better modelling of spatio-temporal dependency in multimodal time-series sensor signal. These findings have significant implications for the understanding of how to construct feature representation that leverages better separability for classification and unseen class detection. The findings reported here lays the groundwork of future research into natural language description generation from multimodal sensor signals. However, publicly available benchmark HAR dataset is limited by the lack of complex activity annotations. Notwithstanding these limitations, the hierarchical self-attention model demonstrates interpretable activity recognition as well as robust feature representation.

\subsection*{Acknowledgments}
This project is supported by ICT Division, Government of Bangladesh, and Independent University, Bangladesh (IUB).
%
%
%
\bibliographystyle{splncs04}
\bibliography{pakdd21}
%




\end{document}